\documentclass[journal]{IEEEtran}
\usepackage{csquotes}
\usepackage{graphicx}
\usepackage{soul}
\usepackage{float}
\usepackage{multirow}
\usepackage{booktabs}
\DeclareUnicodeCharacter{0301}{\'{e}}
\usepackage{amsmath}
\usepackage{amssymb}
\usepackage{amsfonts}
\usepackage[table]{xcolor}
\usepackage[normalem]{ulem}
\usepackage{xspace}
\usepackage{makecell}
\usepackage{tabularx}   % in your preamble

\usepackage{subcaption}

\usepackage[margin=0.5in]{geometry}
\usepackage{tikz}
\usepackage{fontawesome5}
\usetikzlibrary{mindmap, shadows, backgrounds}

\usepackage[margin=0.5in]{geometry}
\usepackage{tikz}
\usepackage{fontawesome5}
\usetikzlibrary{mindmap,trees,shadows,backgrounds}

\definecolor{sceneColor}{HTML}{2874A6}        % Blue
\definecolor{activityColor}{HTML}{138D75}     % Teal
\definecolor{genColor}{HTML}{7D3C98}          % Violet

\newcommand{\latinphrase}[1]{\textit{#1}}

\newcommand{\eg}{\latinphrase{e.g.}\xspace}

\usepackage[pagebackref=false,breaklinks=true,colorlinks,bookmarks=false]{hyperref}
\title{Progressive Split Mamba: Effective State Space Modelling for Image Restoration}

\author{Mohammed Hassanin,
        Nour Moustafa,
        Weijian Deng,
        Ibrahim Radwan
       
\thanks{}% <-this % stops a space
\thanks{}% <-this % stops a space
\thanks{}
}
\begin{document}
\maketitle
\begin{abstract}
Image restoration requires simultaneously preserving fine-grained local structures and maintaining long-range spatial coherence. While convolutional networks struggle with limited receptive fields, and Transformers incur quadratic complexity for global attention, recent State Space Models (SSMs), such as Mamba, provide an appealing linear-time alternative for long-range dependency modelling. However, naively extending Mamba to 2D images exposes two intrinsic shortcomings. First, flattening 2D feature maps into 1D sequences disrupts spatial topology, leading to \emph{locality distortion} that hampers precise structural recovery. Second, the stability-driven recurrent dynamics of SSMs induce \emph{long-range decay}, progressively attenuating information across distant spatial positions and weakening global consistency. Together, these effects limit the effectiveness of state-space modelling in high-fidelity restoration.

We propose Progressive Split-Mamba (PS-Mamba), a topology-aware hierarchical state-space framework designed to reconcile locality preservation with efficient global propagation. Instead of sequentially flattening entire feature maps, PS-Mamba performs geometry-consistent partitioning, maintaining neighbourhood integrity prior to state-space processing. A progressive split hierarchy (halves, quadrants, octants) enables structured multi-scale modelling while retaining linear complexity. To counteract long-range decay, we introduce symmetric cross-scale shortcut pathways that directly transmit low-frequency global context across hierarchical levels, stabilising information flow over large spatial extents.

Extensive experiments on super-resolution, denoising, and JPEG artifact reduction show consistent improvements over recent Mamba-based and attention-based models with a clear margin.% Our code and models are available at: \url{http://github.com/mfawzy/PS-Mamba}
% \wj{
% State space models such as Mamba provide an efficient alternative to attention for modeling long-range dependencies. However, their direct application to image restoration reveals a fundamental mismatch: images are inherently two-dimensional, while state-space models operate on one-dimensional sequences. Flattening a feature map into a long scanline disrupts spatial adjacency, forcing local interactions to be modeled through extended recurrent transitions. Meanwhile, information propagated across long state-space chains progressively attenuates, weakening global structural consistency. These limitations arise from the misalignment between 1D sequence modeling and 2D spatial structure.
% %%
% We propose Progressive Split-Mamba (PS-Mamba), a topology-consistent reformulation of state-space modeling for image restoration. Instead of processing the entire feature map as a single sequence, PS-Mamba progressively partitions features into geometry-aligned regions and performs state-space propagation within each region, preserving neighborhood continuity while maintaining linear complexity. In addition, symmetric cross-scale pathways shorten effective propagation depth, stabilizing long-range information flow.
% %%
% Extensive experiments on super-resolution, denoising, and JPEG artifact reduction show consistent improvements over recent Mamba-based and attention-based models under comparable settings.
% }
\end{abstract}
\section{Introduction}
\label{sec:intro}
\begin{figure}[ht]
\centering
%
% ------------------------
% Row 1 (two columns)
% ------------------------
\begin{subfigure}[t]{0.46\linewidth}
    \centering
    \includegraphics[width=\linewidth]{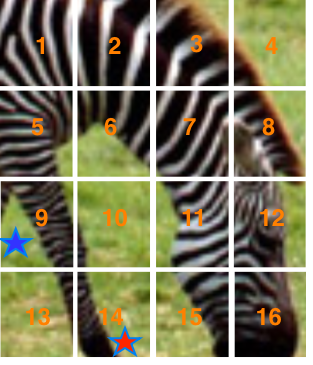}
    \caption{}
\end{subfigure}
% \hfill
\begin{subfigure}[t]{0.5\linewidth}
    \centering
    \includegraphics[width=\linewidth]{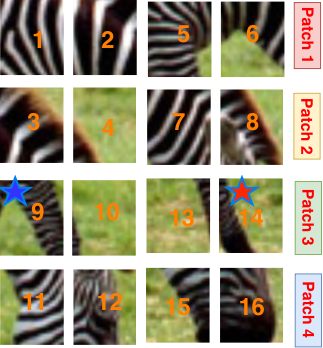}
    \caption{}
\end{subfigure}
% ------------------------
% Row 2 (full width)
% ------------------------
\begin{subfigure}[t]{\linewidth}
    \centering
    \includegraphics[width=\linewidth]{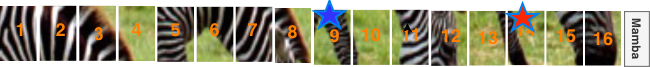}
    \caption{}
\end{subfigure}
% ------------------------
% Row 3 (full width)
% ------------------------
\begin{subfigure}[t]{\linewidth}
    \centering
    \includegraphics[width=\linewidth]{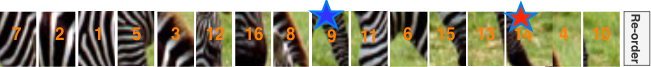}
    \caption{}
\end{subfigure}
\caption{Motivation of using PS-Mamba compared to basic Mamba and recent methods in image restoration. 
(a) The input image is conceptually split into pixel groups (shown enlarged for illustration). 
(b) PS-Mamba processes these geometry-aligned patches independently, preserving neighbourhood proximity and retaining local structure. 
This is contrasted with (c) the basic Mamba formulation, which flattens the entire image into a single long sequence and disrupts spatial adjacency, and (d) re-ordering strategies that require an additional learning stage to embed local textures into a closer sequence space, as in \cite{Guo2025MambaIRv2}. The stars represent the query (red) and reference (blue) pixels for the sequence modelling.}
\label{fig:motivation}
\end{figure}
Image restoration aims to reconstruct high-fidelity images from their degraded inputs, which may suffer from noise, blur, or other distortions. This constitutes an ill-posed problem in computer vision, which includes various sub-tasks such as super-resolution, denoising, and deblurring. Despite recent advances achieved by deep learning methods, particularly Convolutional Neural Networks (CNNs)~\cite{zhang2017learning, ayyoubzadeh2021high, soh2022variational} and Transformers~\cite{zamir2022restormer, wang2022uformer, zhao2023comprehensive, liang2024vrt}, this often comes at the cost of several issues. More precisely, CNNs struggle to capture long-range relationships due to their sensitivity to local receptive fields, whereas Transformers overlook local neighbourhoods due to their global-based attention nature \cite{hassanin2024visual, vaswani2017attention}. Furthermore, CNNs are spatially invariant, making them less responsive to content-specific degradations, while Transformers lack the spatial inductive biases necessary to robustly preserve local structural details. Overcoming these limitations requires a flexible architecture that can simultaneously model fine-grained locality and long-range dependencies.
% \begin{figure}[!ht]
%     \centering
%     \includegraphics[width=\linewidth]{figures/motivation_v3.png}
%     \caption{Caption}
%     \label{fig:placeholder}
% \end{figure}
%

Recently, State Space Models (SSMs) were introduced as a scalable alternative to CNNs and Transformers, which incur significant computational overhead when processing long sequences~\cite{gu2024mamba}. SSMs model dependencies in linear time, making them substantially more efficient. Building on this principle, the Mamba architecture incorporates a selective state-space mechanism that captures long-range interactions while maintaining this efficiency advantage. Extending this further, Mamba-Out adapts SSMs to 2D by applying them directly in the image plane, enabling the model to capture long-range spatial relationships more effectively~\cite{liu2024vmamba}. These properties make Mamba-based models appealing for image restoration, as they strike a balance between global receptive fields and computational efficiency. However, they still exhibit limitations in capturing fine spatial localisation due to their weaker modelling of locality, which can lead to suboptimal performance in dense prediction tasks such as image restoration.
In this work, we introduce \textit{Progressive Split-Mamba (PS-Mamba)}; a framework designed to overcome the two inherent limitations of applying the state-space models to 2D image restoration: locality distortion and long-range decay. Since rasterising the 2D features into 1D sequence disrupts spatial adjacency, Mamba is forced to model local textures through long recurrence chains, while its stability constraint further causes distant information to fade exponentially, as shown in Fig. \ref{fig:motivation}. 
\begin{figure*}[!t]
  \centering
  \includegraphics[width=\textwidth]{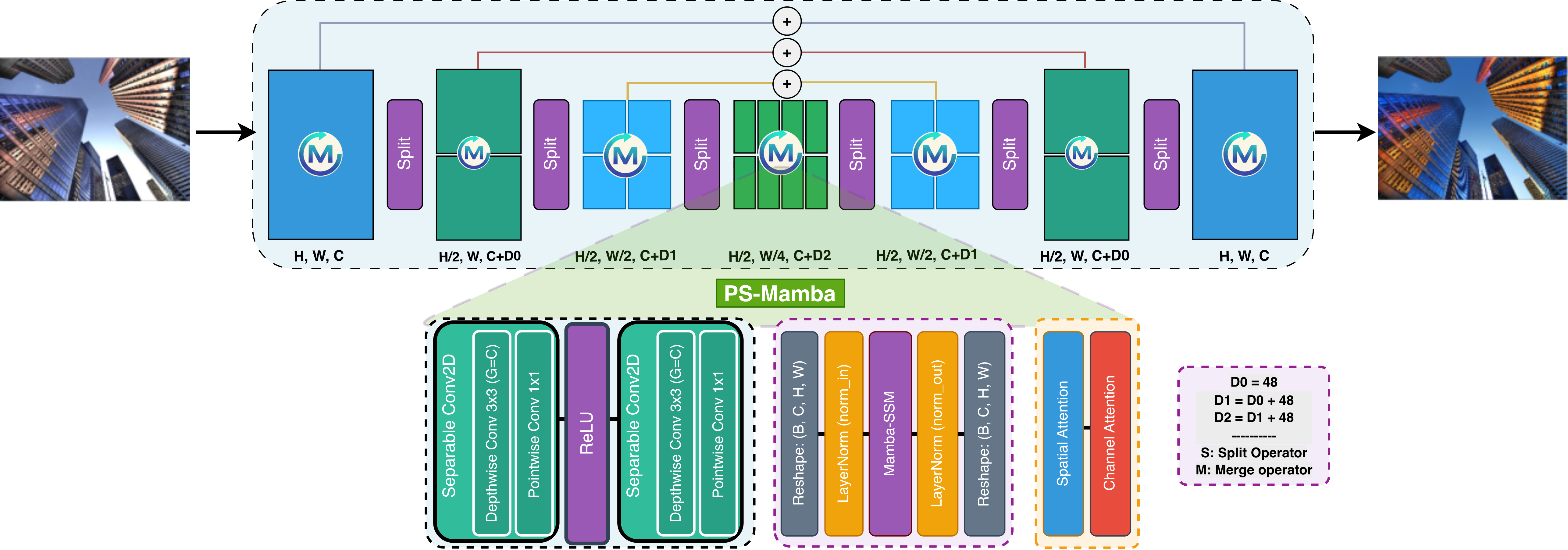}
\caption{Pipeline of the proposed PS-Mamba framework: 
The input image is progressively split into multi-scale, geometry-aligned regions(\eg halves, quadrants, and octants) while maintaining the original spatial resolution. 
These patches enable patch-level processing that preserves locality and mitigates long-range decay. The channel dimension is increased by 48 to enhance representation capacity, and it is symmetrically reduced by 48 during the merging stages to maintain balanced skip connections.}
  
  \label{fig:pipeline}
\end{figure*}
Our key idea is to preserve the geometric locality before state-space propagation by partitioning the image into topology-consistent patches and applying Mamba within each patch. This restores natural pixel neighbourhoods in the induced sequence, allowing the model to capture fine-scale structures without relying on ad-hoc token manipulations or multi-directional scanning. Furthermore, we introduce symmetric cross-scale skip propagation module that delivers global low-frequency context across the network through direct shortcuts, effectively collapsing the long dependency distances responsible for the state-space decay.

Generally, the proposed method makes three central contributions. First, we propose a Progressive Split-based State-space Module, which processes feature maps through a progressive hierarchy of halves, quadrants, and octants of the inputs, while preserving their 2D spatial structure and enabling a multi-layered reasoning within the linear-time constraints of the SSMs. This eliminates raster-induced adjacency errors and ensures stable modelling of fine textures. Second, we introduce symmetric cross-scale skip links that bypass the long Mamba chain, delivering global structural information directly between mirrored stages. These skip links counteract the exponential decay in linear SSMs, and reinforce low-frequency consistency and the stability of gradient flow. Third, by integrating the proposed SSM with an adaptive Mamba-based Convolution blocks and dual attention refinement, we develop a restoration architecture that unifies the strengths of state-space modelling and locality consideration. Extensive experiments are performed to ensure that PS-Mamba provides significant gains in both detail fidelity and global coherence relevant to the image restoration tasks.
\section{Related Work}
\label{sec::related}
% This section reviews previous work in image restoration across three main families of architectures: CNN-based models, Transformer-based models, and the recent Vision Mamba based approaches. We highlight the strengths of each category and the key limitations that motivate the development of the proposed method.
%
\paragraph{CNN-based methods:}
Following the success of CNNs in visual recognition, early restoration networks such as SRCNN~\cite{dong2016image} for super-resolution, DnCNN~\cite{zhang2017beyond} for denoising, and ARCNN~\cite{dong2015compression} for JPEG artefact reduction demonstrated the effectiveness of convolutional priors in handling local degradations. Inspired by deeper architectures such as ResNet and DenseNet~\cite{huang2017densely, he2016deep}, subsequent work introduced residual connections and attention modules~\cite{zhang2018image,  dai2019second}, enabling more expressive CNNs capable of modelling richer spatial patterns.  
However, these approaches remain inherently limited by their locality, \textit{i.e.,} their fixed receptive fields make it difficult to capture long-range dependencies, which are necessary in the image restoration tasks.

\paragraph{Transformer-based methods:}
Transformers~\cite{vaswani2017attention, dosovitskiy2020image} brought a major shift in image restoration due to their ability to encode global relationships through self-attention. Numerous work~\cite{chen2021pre, liang2021swinir, chen2023hat} adapted attention mechanisms to restoration tasks, achieving strong performance by modelling non-local interactions.  
For example, SwinIR~\cite{liang2021swinir} used shifted-window attention to balance efficiency and global context, while ART~\cite{zhang2022accurate}  employed sparse and structured attention to enlarge the receptive field. Despite these advantages, Transformer-based designs encounter two main issues: (1) high computational cost, particularly for high-resolution inputs; and (2) loss of local fidelity due to window partitioning or globally focused attention, which may weaken fine-scale structures.

\paragraph{Vision Mamba:}
The recently proposed Mamba architecture~\cite{Guo2024MambaIR} offers an appealing alternative by modelling long-range dependencies through selective state-space mechanisms in linear time. This efficiency has led to a growing interest in applying Mamba to vision tasks such as image recognition~\cite{liu2024vmamba}.  
MambaIR~\cite{Guo2024MambaIR} introduced Mamba blocks for restoration, while FreqMamba~\cite{zou2024freqmamba} incorporated frequency-domain state modelling. For de-raining, MambaLLIE~\cite{weng2024mamballie} improved locality in low-light enhancement. MambaIRv2~\cite{Guo2025MambaIRv2} extended the approach using shifted windows and visual prompt pooling to further refine spatial reasoning. Recent studies have explored adapting Mamba-based state space models to image restoration by introducing locality-aware designs. 
LocalMamba~\cite{huang2024localmamba} restricts the Mamba scanning operation to fixed spatial windows in order to reduce sequence length and improve local feature modeling. 
Similarly, MaIR~\cite{li2025mair} introduces locality and continuity constraints to better preserve spatial structures when applying Mamba to image restoration tasks. 
Other works, such as VmambaIR, extend the visual state space model framework to restoration problems by applying Mamba modules over global image representations. Despite these advancements, current Mamba-based restoration methods face two fundamental limitations:  
(1) \emph{locality distortion}, where flattening 2D features into a 1D sequence forces the model to prioritise long-range transitions over true local neighbourhoods; and  
(2) \emph{long-range decay}, where the stability constraints of linear state-space propagation cause distant information to fade as the sequence length increases. To address these limitations, our proposed \textbf{PS-Mamba} introduces a hierarchical split-and-merge strategy that preserves true 2D adjacency without requiring any token shifting or reordering. This design maintains geometric locality before state-space propagation and avoids disrupting the inherent spatial structure of the features. Additionally, symmetric skip connections mitigate long-range decay by providing direct global pathways, ensuring stable gradient flow and consistent structural information across all stages.  
This formulation offers a general and computation-neutral solution for enhancing locality and global consistency in Mamba-based image restoration. Further, different from previous local-mamba methods, PS-Mamba effectively processes feature maps in multi-scale progressive splits which provide a better balance between modeling capacity and computational efficiency for image restoration tasks.
\vspace{-10pt}
\section{Method}
\label{sec:method}
% \paragraph{State Space Model (SSM) foundation} 
In this section, we introduce the proposed \textit{Progressive Split-based Mamba (PS-Mamba)} framework, which is designed to resolve two fundamental limitations of applying basic state-space models such as Mamba to 2D image restoration: (\textit{i}) the {locality distortion} that emerges when 2D feature maps are rasterised into 1D token sequences, and (\textit{ii}) the {long-range decay} inherent to linear state-space propagation, which prevents distant visual dependencies from being preserved during deep processing. We first formalise these challenges, and then present a sequence of architectural components that collectively address them. An overview of the full architecture is shown in Fig.~\ref{fig:pipeline}.

\subsection{Limitations of State Space Module for Image Restoration}
Recently, Mamba modules have been utilised widely in visual applications and restoration for their ability to encode long-range relationships. They achieved progress in efficiency and effectiveness due to their linear time processing. We begin by recalling the standard linear state-space formulation used in Mamba, expressed as:
\begin{equation}
    \mathbf{h}_i = A\mathbf{h}_{i-1} + B\mathbf{x}_i,\qquad
    \mathbf{y}_i = C\mathbf{h}_i + D\mathbf{x}_i,
\label{eq:ssm}
\end{equation}
where $\mathbf{x}_i$ and $\mathbf{h}_i$ denote the input and hidden state at step $i$, $\mathbf{y}_i$ is the output, and $A$, $B$, $C$, and $D$ are the learnable state transition, input, output, and skip matrices, respectively.

Although the basic Mamba state-space formulation is highly effective for 1D sequence modelling due to its linear-time processing and ability to encode long-range dependencies, directly applying it to 2D images introduces structural limitations. When an image feature tensor $\mathbf{X}\in\mathbb{R}^{H\times W\times C}$ is flattened into a 1D sequence of length $L=HW$, naturally 4-connected neighbouring pixels may become separated by $\mathcal{O}(W)$ positions. This flattening distorts locality and forces Mamba to model fine-grained textures through unnecessarily long-range transitions instead of short-range spatial interactions.

Several attempts, such as bidirectional scanning, window reordering, or token rearrangement, partially reduce these effects but remain constrained because they manipulate the input rather than addressing the underlying cause. Moreover, these manipulations are not neighbour-preserving, further disrupting spatial structure. In addition, stability in Mamba often requires the transition matrix to have spectral radius below one, resulting in exponential decay of information as it propagates along the sequence. For images, where thousands of recurrent steps arise from scanning full-resolution inputs, global and low-frequency signals weaken before reaching the relevant positions. Hence, both locality distortion and long-range decay are structural consequences of applying a 1D state-space sequence to inherently 2D data, and resolving only one of them leads to a trade-off between strong local details and coherent global structure.

\subsection{Progressive Split-based State Space Module (PS-Mamba)}
\label{sec:psmamba}

After identifying the limitations of applying a full 1D state-space sequence to 2D images---namely locality distortion and long-range decay---we introduce the \textbf{Progressive Split-based State Space Module} (PS-Mamba). The goal is to restore spatial coherence while preserving the linear-time advantages of Mamba. We achieve this by (1) operating on geometry-preserving patches, (2) enforcing structural refinement before sequence modelling, and (3) stabilising long-range propagation through adaptive fusion. The whole pipeline is shown in Fig. \ref{fig:pipeline}.

\paragraph{Split and Merge Operators}
We begin by defining two operators that form the backbone of PS-Mamba.  
The \emph{split operator} $\mathcal{S}_k(\cdot)$ partitions a feature map into $k$ contiguous and geometry-aligned regions:
\begin{equation}
\mathcal{S}_k : \mathbb{R}^{H\times W\times C} \rightarrow 
\big\{\mathbf{X}^{(1)}, \dots, \mathbf{X}^{(k)}\big\},
\qquad
\mathbf{X}^{(j)}\in\mathbb{R}^{H_j\times W_j\times C}.
\end{equation}
Each $\mathbf{X}^{(j)}$ is reshaped into a sequence of length $L_j = H_jW_j$ while preserving 4-connected adjacency.  
After patch-wise processing, the \emph{merge operator} $\mathcal{M}_k(\cdot)$ reconstructs the full spatial layout:
\begin{equation}
\mathbf{Y} = 
\mathcal{M}_k\Big(
\mathrm{SSM}(\mathrm{Seq}(\mathbf{X}^{(1)})), \dots,
\mathrm{SSM}(\mathrm{Seq}(\mathbf{X}^{(k)}))
\Big).
\end{equation}
Since $L_j \ll HW$, the recurrent depth per patch is short, providing stable and locality-preserving state-space modelling.

\paragraph{Convolutional Preprocessing}
Before feeding features into Mamba, we apply a lightweight convolutional refinement to enforce local continuity and enhance structural patterns.  
Given input $x\in\mathbb{R}^{B\times C\times H\times W}$, we extract short-range textures through a residual convolution branch:
\begin{equation}
\mathbf{f}_{\mathrm{conv}} = 
\mathrm{Conv}_{3\times 3}\!\left(
\mathrm{ReLU}\big(\mathrm{Conv}_{3\times 3}(x)\big)
\right).
\end{equation}
This step ensures that Mamba receives already stabilised, edge-aware features rather than raw unstructured activations, reducing the burden on the sequence model and preventing early loss of local detail.

\paragraph{Patch-level Mamba Core}
After convolutional refinement, each patch $\mathbf{X}^{(j)}$ is processed independently through the Mamba sequence:
\begin{equation}
\mathbf{h}^{(j)}_{i} = A\mathbf{h}^{(j)}_{i-1} + B\mathbf{x}^{(j)}_{i}, 
\qquad
\mathbf{y}^{(j)}_{i} = C\mathbf{h}^{(j)}_{i} + D\mathbf{x}^{(j)}_{i},
\label{eq:patchcore}
\end{equation}
for $i=1\ldots L_j$.  
A pair of normalisation layers stabilises this patch-wise recurrent computation:
\begin{equation}
\mathbf{f}_{\mathrm{m}} = 
\mathrm{Fold}\!\left(
\mathrm{LN}_{\mathrm{out}}\!\left(
\mathrm{Mamba}\big(\mathrm{LN}_{\mathrm{in}}(\mathrm{Unfold}(x))\big)
\right)\right).
\end{equation}
Processing within patches eliminates adjacency violations and prevents the exponential attenuation that occurs when Mamba is applied to a single very long sequence.

\paragraph{Attention-based Fusion}
Although patch-level Mamba restores locality, long-range decay still appears if global signals are not reinforced.  
To counter this, we fuse convolutional and Mamba features using content-adaptive gating:
\begin{align}
\mathbf{g} &= 
\sigma\!\big(
W_2\,\delta(W_1\,\mathrm{GAP}(\mathbf{f}_{\mathrm{conv}} + \mathbf{f}_{\mathrm{m}}))
\big),\\
\mathbf{f}_{\mathrm{mix}} &= 
\mathbf{g}\odot \mathbf{f}_{\mathrm{conv}} +
(1{-}\mathbf{g})\odot \mathbf{f}_{\mathrm{m}}.
\end{align}
where $W_1$ and $W_2$ are the learnable weight matrices of two-layer MLP, $\mathrm{GAP}$ is the Global Average Pooling that reduces the spatial map to a channel vector of size $C$, $\delta$ is ReLU, and $\sigma$ is a sigmoid function that produces a channel-wise gating vector.

Then this is followed by a dual attention refinement; channel attention (CA) and spatial attention (SA). This further emphasises discriminative structures and reinforces global consistency as follows:
\begin{equation}
\mathbf{y} = x + \alpha \cdot \mathrm{SA}\!\left(\mathrm{CA}(\mathbf{f}_{\mathrm{mix}})\right)
\end{equation}
where $\alpha$ controls how much of the refined feature map is added to the input via the residual connection.  This fusion step strengthens low-frequency information, stabilises long-range propagation, and complements the locality captured within patches.

Overall PS-Mamba integrates  
(1) a convolutional path for structural priors,  
(2) a topology-preserving patch-level Mamba sequence, and  
(3) an attention-based fusion block that mitigates long-range decay.  
Together, these components restore locality, preserve efficiency, and produce significantly improved spatial fidelity compared to naive full-sequence Mamba.
\subsection{Training Objective and Optimisation}
\label{sec:loss}
Unlike previous progressive restoration networks~\cite{zamir2021multi}, which assign independent losses to each intermediate stage, our method employs a single, unified $L_1$ loss that jointly supervises both the final and stage-wise predictions. This design simplifies optimisation and encourages consistent feature refinement across all stages without requiring separate objectives.
%

%
%%%%%%%%%%%%%%%

\paragraph{Loss Function}
For image restoration, we adopt the pixel-wise $L_1$ loss to optimise the parameters of the network, following the previous research attempts~\cite{liang2021swinir, mambair, srformer}. The objective function is defined as
\begin{equation}
\mathcal{L} = \|y - \hat{y}\|_1,
\end{equation}
where $y$ denotes the ground-truth image and $\hat{y}$ is the predicted output.
Unlike previous progressive restoration frameworks that apply separate supervision to each stage, our method employs a single unified $L_1$ loss over the final output, ensuring stable and efficient convergence while reducing optimisation redundancy.
For other restoration tasks such as image denoising, we utilise the Charbonnier loss~\cite{charbonnier1994two}, a differentiable variant of the $L_1$ norm, defined as
\begin{equation}
\mathcal{L} = \sqrt{\|y - \hat{y}\|_2^2 + \epsilon^2},
\end{equation}
where $\epsilon$ is a small constant empirically set to $10^{-3}$ to enhance numerical stability.

\paragraph{Complexity Analysis.}
Let the input feature map be $\mathbf{X} \in \mathbb{R}^{H \times W \times C}$. 
In the standard Mamba formulation, the 2D feature map is flattened into a sequence 
of length $L = HW$. Consequently, the state-space module processes a single long 
sequence whose recurrent depth grows proportionally with the spatial resolution.
In contrast, PS-Mamba partitions the feature map into $k$ spatially contiguous 
regions of size $H_j \times W_j$, such that $\sum_{j=1}^{k} H_j W_j = HW$. 
Each region is processed independently through the state-space module, producing 
sequences of length $L_j = H_j W_j \ll HW$.

Therefore, the total computational complexity of PS-Mamba remains

\[
\mathcal{O}\left(\sum_{j=1}^{k} L_j \right) = \mathcal{O}(HW),
\]
which preserves the linear complexity property of state-space models. 
However, the effective recurrent depth for each sequence is significantly reduced, 
leading to more stable information propagation and improved preservation of local 
spatial structures.
\begin{table*}[ht]
\centering
\caption{Quantitative comparison on \textit{lightweight image super-resolution} with state-of-the-art methods. The best and the second best results are in \textcolor{red}{red} and \textcolor{blue}{blue}.}
\resizebox{\textwidth}{!}{
\begin{tabular}{l|c|c|c|cc|cc|cc|cc|cc}
\toprule
\multirow{2}{*}{Method} & \multirow{2}{*}{scale} & \multirow{2}{*}{\#param} & \multirow{2}{*}{MACs} &
\multicolumn{2}{c}{Set5} & \multicolumn{2}{c}{Set14} & \multicolumn{2}{c}{BSDS100} & \multicolumn{2}{c}{Urban100} & \multicolumn{2}{c}{Manga109} \\
\cmidrule(lr){5-6} \cmidrule(lr){7-8} \cmidrule(lr){9-10} \cmidrule(lr){11-12} \cmidrule(lr){13-14}
 & & & & PSNR & SSIM & PSNR & SSIM & PSNR & SSIM & PSNR & SSIM & PSNR & SSIM \\
\midrule
CARN~\cite{ahn2018fast} & 2$\times$ & 1,592K & 222.8G & 37.76 & 0.9590 & 33.52 & 0.9166 & 32.09 & 0.8978 & 31.92 & 0.9256 & 38.36 & 0.9765 \\
LatticeNet~\cite{luo2020latticenet} & 2$\times$ & 756K & 169.5G & 38.13 & 0.9610 & 33.78 & 0.9193 & 32.25 & 0.9005 & 32.43 & 0.9302 & - & - \\
SwinIR-light~\cite{liang2021swinir} & 2$\times$ & 910K & 244.2G & 38.14 & 0.9611 & 33.86 & 0.9206 & 32.31 & 0.9012 & 32.76 & 0.9340 & 39.12 & 0.9783 \\
MambaIR-light~\cite{mambair} & 2$\times$ & 905K & 334.2G &  38.13& 0.9610 &33.95& 0.9208& 32.31& 0.9013 &32.85 &0.9349& 39.20& 0.9782\\
ELAN~\cite{zhang2021efficient} & 2$\times$ & 621K & 203.1G & 38.17 & 0.9611 & 33.94 & 0.9207 & 32.30 & 0.9012 & 32.76 & 0.9340 & 39.11 & 0.9782 \\
SRFormer-light~\cite{srformer} & 2$\times$ & 853K & 236.3G & 38.23 & 0.9613 & 33.94 & 0.9209 & 32.36 & 0.9019 & 32.91 & 0.9353 & 39.28 & 0.9785 \\
MaIR-Small \cite{mair} &2$\times$&1,355K& 542.0G& 38.20& 0.9611& 33.91& 0.9209 &32.34& 0.9016& 32.97& 0.9359 &39.32 &0.9779 \\
%
%
%%\rowcolor{blue!10}
MambaIRv2-light \cite{Guo2025MambaIRv2}& 2 $\times$ &774K & 126.7G &
\textcolor{blue}{38.26} & \textcolor{blue}{0.9615} & 
\textcolor{blue}{34.09} & \textcolor{blue}{0.9221} & 
\textcolor{blue}{32.36} & \textcolor{blue}{0.9019} & 
\textcolor{blue}{33.26} & \textcolor{blue}{0.9378} & 
\textcolor{blue}{39.35} & \textcolor{blue}{0.9785} \\
%
%%\rowcolor{red!10}
PS-Mamba light (Ours) &2 $\times$&970K&60.1G& 
\textcolor{red}{38.31} & \textcolor{red}{0.9618} & 
\textcolor{red}{34.38} & \textcolor{red}{0.9238} & 
\textcolor{red}{32.43} & \textcolor{red}{0.9029} & 
\textcolor{red}{33.37} & \textcolor{red}{0.9394} & 
\textcolor{red}{39.74} & \textcolor{red}{0.9795} \\
\midrule
CARN~\cite{ahn2018fast} & 3$\times$ & 1,592K & 118.8G & 34.29 & 0.9255 & 30.29 & 0.8407 & 29.06 & 0.8034 & 28.06 & 0.8493 & 33.50 & 0.9440 \\
LatticeNet~\cite{luo2020latticenet} & 3$\times$ & 765K & 76.3G & 34.53 & 0.9281 & 30.39 & 0.8424 & 29.15 & 0.8059 & 28.33 & 0.8538 & - & - \\
SwinIR-light~\cite{liang2021swinir} & 3$\times$ & 918K & 111.2G & 34.62 & 0.9290 & 30.54 & 0.8463 & 29.22 & 0.8082 & 28.66 & 0.8624 & 33.98 & 0.9478 \\
MambaIR-light~\cite{mambair} & 3$\times$ & 913K & 148.5G & 34.61 & 0.9288 & 30.54 & 0.8459 & 29.23 & 0.8084 & 28.70 & 0.8631 & 34.12 & 0.9497 \\
ELAN~\cite{zhang2021efficient} & 3$\times$ & 629K & 90.1G & 34.61 & 0.9290 & 30.58 & 0.8469 & 29.21 & 0.8081 & 28.69 & 0.8624 & 34.00 & 0.9489 \\
SRFormer-light~\cite{srformer} & 3$\times$ & 861K & 105.4G & 34.67 & 0.9296 & 30.57 & 0.8469 & 29.26 & 0.8099 & 28.81 & 0.8655 & 34.19 & 0.9489 \\
MaIR-Small \cite{mair}&3$\times$&1,363K& 241.4G& 34.75 &0.9300& 30.63 &0.8479 &29.29& 0.8103 &28.92& 0.8676& 34.46& 0.9497 \\
%
%\rowcolor{blue!10}
MambaIRv2-light \cite{Guo2025MambaIRv2} & 3$\times$ & 781K &126.7G & \textcolor{blue}{34.71} & \textcolor{blue}{0.9298} & \textcolor{blue}{30.68} & \textcolor{blue}{0.8483} & \textcolor{blue}{29.26} & \textcolor{blue}{0.8098} & \textcolor{blue}{29.01} & \textcolor{blue}{0.8689} & \textcolor{blue}{34.41} & \textcolor{blue}{0.9497} \\
%\rowcolor{red!10}
PS-Mamba-light (Our) & 3$\times$ &  970K&60.1G&\textcolor{red}{34.84}  &\textcolor{red}{0.9304} &\textcolor{red}{30.76}  &\textcolor{red}{0.8494} &  \textcolor{red}{29.35} &\textcolor{red}{0.8118}  & \textcolor{red}{29.26} & \textcolor{red}{0.8731}& \textcolor{red}{34.84} &\textcolor{red}{0.9516}  \\
\midrule
CARN~\cite{ahn2018fast} & 4$\times$ & 1,592K & 90.8G & 32.13 & 0.8937 & 28.60 & 0.7806 & 27.58 & 0.7367 & 26.07 & 0.7837 & 30.47 & 0.9084 \\
LatticeNet~\cite{luo2020latticenet} & 4$\times$ & 765K & 43.6G & 32.00 & 0.8960 & 28.68 & 0.7830 & 27.63 & 0.7376 & 26.23 & 0.7873 & - & - \\
SwinIR-light~\cite{liang2021swinir} & 4$\times$ & 912K & 63.6G & 32.44 & 0.8976 & 28.83 & 0.7877 & 27.69 & 0.7404 & 26.47 & 0.7982 & 30.92 & 0.9151 \\
MambaIR-light~\cite{mambair} & 4$\times$ & 924K & 84.6G & 32.42 & 0.8975 & 28.82 & 0.7878 & 27.70 & 0.7406 & 26.54 & 0.7992 & 31.06 & 0.9172 \\
ELAN~\cite{zhang2021efficient} & 4$\times$ & 640K & 52.1G & 32.43 & 0.8976 & 28.82 & 0.7878 & 27.70 & 0.7404 & 26.54 & 0.7992 & 31.04 & 0.9170 \\
SRFormer-light~\cite{srformer} & 4$\times$ & 873K & 62.8G &
32.51 & 0.8998 & 28.84 & 0.7878 & 27.75 & 0.7426 &
26.67 & 0.8032 & 31.17 & 0.9165 \\
MaIR-Small \cite{mair} &4$\times$&1,374K &136.6G& 32.62& 0.8998& 28.90& 0.7882& 27.77& 0.7431& 26.73& 0.8049& 31.34& 0.9183 \\
%\rowcolor{blue!10}
MambaIRv2-light \cite{Guo2025MambaIRv2}  & 4$\times$ & 790K & 75.6G & 
\textcolor{blue}{32.51} & \textcolor{blue}{0.8992} & 
\textcolor{blue}{28.84} & \textcolor{blue}{0.7878} & 
\textcolor{blue}{27.75} & \textcolor{blue}{0.7426} & 
\textcolor{blue}{26.82} & \textcolor{blue}{0.8079} & 
\textcolor{blue}{31.24} & \textcolor{blue}{0.9182} \\
%
%\rowcolor{red!10}
PS-Mamba(Ours) & 4$\times$ & 970K&60.1G&
\textcolor{red}{32.71} & \textcolor{red}{0.9006} &
\textcolor{red}{28.96} & \textcolor{red}{0.7902} &
\textcolor{red}{27.82} & \textcolor{red}{0.7442} &
\textcolor{red}{27.02} & \textcolor{red}{0.8127} &
\textcolor{red}{31.59} & \textcolor{red}{0.9202} \\
\bottomrule
\end{tabular}
\label{tab:light}
}
\end{table*}
\begin{table*}
\centering
\caption{Quantitative comparison on \textit{classic image super-resolution} with state-of-the-art methods. The best and the second best results are in \textcolor{red}{red} and \textcolor{blue}{blue}.}
\resizebox{\textwidth}{!}{
\begin{tabular}{l|c|c|cc|cc|cc|cc|cc}
\toprule
\multirow{2}{*}{Method} & \multirow{2}{*}{scale} & \multirow{2}{*}{\#param} &
\multicolumn{2}{c}{Set5} & \multicolumn{2}{c}{Set14} & \multicolumn{2}{c}{BSDS100} &
\multicolumn{2}{c}{Urban100} & \multicolumn{2}{c}{Manga109} \\
\cmidrule(lr){4-5}\cmidrule(lr){6-7}\cmidrule(lr){8-9}\cmidrule(lr){10-11}\cmidrule(lr){12-13}
& & & PSNR & SSIM & PSNR & SSIM & PSNR & SSIM & PSNR & SSIM & PSNR & SSIM \\
\midrule
EDSR~\cite{lim2017enhanced} & 2$\times$ & 42.6M & 38.20 & 0.9613 & 33.92 & 0.9195 & 32.32 & 0.9013 & 32.93 & 0.9351 & 39.10 & 0.9773 \\
RCAN~\cite{zhang2018image} & 2$\times$ & 15.4M & 38.27 & 0.9614 & 34.12 & 0.9216 & 32.41 & 0.9027 & 33.34 & 0.9384 & 39.44 & 0.9786 \\
SAN~\cite{dai2019second} & 2$\times$ & 15.7M & 38.31 & 0.9620 & 34.07 & 0.9213 & 32.42 & 0.9028 & 33.10 & 0.9370 & 39.32 & 0.9792 \\
IPT~\cite{chen2021pre} & 2$\times$ & 115M & 38.37 & 0.9620 & 34.43 & 0.9250 & 32.48 & 0.9038 & 33.76 & 0.9404 & 39.72 & 0.9789 \\
SwinIR~\cite{liang2021swinir} & 2$\times$ & 11.8M & 38.42 & 0.9623 & 34.46 & 0.9250 & 32.53 & 0.9041 & 33.81 & 0.9427 & 39.60 & 0.9785 \\
EDT~\cite{edt2023} & 2$\times$ & 11.5M & 38.45 & 0.9624 & 34.47 & 0.9253 & 32.54 & 0.9044 & 33.85 & 0.9441 & 39.93 & 0.9800 \\
MambaIR~\cite{mambair} & 2$\times$ & 20.4M & 38.57 & 0.9627 & 34.70 & 0.9267 & 32.61 & 0.9054 & 34.04 & 0.9446 & 40.08 & 0.9806 \\
CAT-A~\cite{cata2023} & 2$\times$ & 16.5M & 38.56 & 0.9627 & 34.67 & 0.9263 & 32.60 & 0.9053 & 34.01 & 0.9444 & 40.06 & 0.9803 \\
DAT~\cite{dat2023} & 2$\times$ & 14.8M & 38.56 & 0.9627 & 34.78 & 0.9276 & 32.60 & 0.9053 & 34.25 & 0.9468 & 40.32 & 0.9808 \\
HAT~\cite{hat2023} & 2$\times$ & 20.6M & 38.63 & 0.9630 & 34.86 & 0.9274 & 32.62 & 0.9053 & 34.45 & 0.9466 & 40.26 & 0.9809 \\
MaIR~\cite{mair} & 2$\times$ & -- & 38.62 &0.9630 &34.82& 0.9272 &32.62& 0.9053 &34.38 &0.9462 &40.48 &\textcolor{blue}{0.9811}\\
% %%\rowcolor{blue!10}
% %%\rowcolor{blue!10}
MambaIRv2-L\cite{Guo2025MambaIRv2}  & 2$\times$ & 34.2M & \textcolor{blue}{38.65} & \textcolor{blue}{0.9631} & \textcolor{blue}{34.93} & \textcolor{blue}{0.9276} & \textcolor{blue}{32.62} & \textcolor{blue}{0.9053} & \textcolor{blue}{34.60} & \textcolor{blue}{0.9475} & \textcolor{blue}{40.55} & \textcolor{black}{0.9807} \\
%%\rowcolor{red!10}

  SP-Mamba-L (Ours) & 2$\times$ & 21.2M & \textcolor{red}{38.70} & \textcolor{red}{0.9633} & \textcolor{red}{35.20} & \textcolor{red}{0.9289} & \textcolor{red}{32.69} & \textcolor{red}{0.9062} & \textcolor{red}{34.71} & \textcolor{red}{0.9486} & \textcolor{red}{40.90} & \textcolor{red}{0.9814} \\

\specialrule{1pt}{0pt}{0pt}
EDSR~\cite{lim2017enhanced} & 3$\times$ & 42.6M & 34.65 & 0.9280 & 30.52 & 0.8462 & 29.25 & 0.8093 & 28.80 & 0.8653 & 34.17 & 0.9476 \\
RCAN~\cite{zhang2018image} & 3$\times$ & 15.4M & 34.74 & 0.9290 & 30.65 & 0.8482 & 29.32 & 0.8111 & 29.09 & 0.8702 & 34.47 & 0.9496 \\
SAN~\cite{dai2019second} & 3$\times$ & 15.7M & 34.75 & 0.9290 & 30.59 & 0.8476 & 29.33 & 0.8112 & 28.93 & 0.8686 & 34.30 & 0.9483 \\
IPT~\cite{chen2021pre} & 3$\times$ & 115M & 35.07 & 0.9300 & 31.07 & 0.8534 & 29.59 & 0.8154 & 29.81 & 0.8830 & 35.12 & 0.9537 \\
SwinIR~\cite{liang2021swinir} & 3$\times$ & 11.8M & 34.97 & 0.9291 & 30.93 & 0.8534 & 29.50 & 0.8145 & 29.68 & 0.8826 & 35.12 & 0.9537 \\
EDT~\cite{edt2023} & 3$\times$ & 11.5M & 34.97 & 0.9310 & 30.94 & 0.8537 & 29.51 & 0.8149 & 29.76 & 0.8841 & 35.43 & 0.9554 \\
MambaIR~\cite{mambair} & 3$\times$ & 20.4M & 35.06 & 0.9310 & 31.13 & 0.8560 & 29.57 & 0.8162 & 30.00 & 0.8886 & 35.35 & 0.9547 \\
CAT-A~\cite{cata2023} & 3$\times$ & 16.5M & 35.06 & 0.9310 & 31.12 & 0.8558 & 29.56 & 0.8161 & 30.00 & 0.8885 & 35.34 & 0.9546 \\
DAT~\cite{dat2023} & 3$\times$ & 14.8M & 35.11 & 0.9310 & 31.18 & 0.8564 & 29.59 & 0.8165 & 30.14 & 0.8905 & 35.53 & 0.9554 \\
HAT~\cite{hat2023} & 3$\times$ & 20.6M & 35.17 & 0.9316 & 31.26 & 0.8574 & 29.61 & 0.8169 & 30.28 & 0.8925 & 35.58 & 0.9559 \\
% %%\rowcolor{blue!10}
MaIR~\cite{mair} & 3$\times$ & -- &35.15 &0.9328 &31.12& 0.8550 &29.56 &0.8167 &30.24 &0.8881 &35.67 &0.9556\\
%%\rowcolor{blue!10}
MambaIRv2-L \cite{Guo2025MambaIRv2}& 3$\times$ & 34.2M & \textcolor{blue}{35.16} & \textcolor{blue}{0.9319} & \textcolor{blue}{31.18} & \textcolor{blue}{0.8564} & \textcolor{blue}{29.57} & \textcolor{blue}{0.8170} & \textcolor{blue}{30.34} & \textcolor{blue}{0.8919} & \textcolor{blue}{35.72} & \textcolor{blue}{0.9561} \\
%%\rowcolor{red!10}
PS-Mamba-L (Ours)  & 3$\times$ & 21.2M & 
\textcolor{red}{35.29} & \textcolor{red}{0.9325}& 
\textcolor{red}{31.26} & \textcolor{red}{0.8575} & 
  \textcolor{red}{29.66} & \textcolor{red}{0.8190} & 
  \textcolor{red}{30.59} & \textcolor{red}{0.8961} & 
  \textcolor{red}{36.15} & \textcolor{red}{0.9580} \\
\specialrule{1pt}{0pt}{0pt}
EDSR~\cite{lim2017enhanced} & 4$\times$ & 43.0M & 32.46 & 0.8968 & 28.80 & 0.7876 & 27.71 & 0.7420 & 26.64 & 0.8031 & 31.02 & 0.9148 \\
RCAN~\cite{zhang2018image} & 4$\times$ & 15.4M & 32.63 & 0.9002 & 28.87 & 0.7889 & 27.77 & 0.7443 & 26.82 & 0.8086 & 31.18 & 0.9198 \\
SAN~\cite{dai2019second} & 4$\times$ & 15.7M & 32.64 & 0.9003 & 28.92 & 0.7892 & 27.78 & 0.7446 & 26.79 & 0.8068 & 31.18 & 0.9198 \\
IPT~\cite{chen2021pre} & 4$\times$ & 115M & 33.02 & 0.9050 & 29.27 & 0.7950 & 27.99 & 0.7490 & 27.52 & 0.8254 & 32.03 & 0.9323 \\
SwinIR~\cite{liang2021swinir} & 4$\times$ & 11.9M & 32.92 & 0.9044 & 29.09 & 0.7917 & 27.92 & 0.7476 & 27.45 & 0.8246 & 32.03 & 0.9323 \\
EDT~\cite{edt2023} & 4$\times$ & 11.5M & 33.02 & 0.9061 & 29.18 & 0.7934 & 27.97 & 0.7488 & 27.69 & 0.8287 & 32.32 & 0.9336 \\
MambaIR~\cite{mambair} & 4$\times$ & 20.4M & 33.03 & 0.9065 & 29.23 & 0.7941 & 27.98 & 0.7495 & 27.83 & 0.8327 & 32.33 & 0.9340 \\
CAT-A~\cite{cata2023} & 4$\times$ & 16.5M & 33.04 & 0.9065 & 29.23 & 0.7940 & 27.98 & 0.7494 & 27.82 & 0.8324 & 32.32 & 0.9337 \\
DAT~\cite{dat2023} & 4$\times$ & 14.8M & 33.09 & 0.9065 & 29.25 & 0.7947 & 28.00 & 0.7503 & 27.90 & 0.8352 & 32.45 & 0.9353 \\
HAT~\cite{hat2023} & 4$\times$ & 20.6M & 33.14 & 0.9072 & 29.29 & 0.7955 & 28.01 & 0.7507 & 28.07 & 0.8384 & 32.57 & 0.9369 \\
MaIR~\cite{mair} & 4$\times$ & -- &33.14 &0.9058 &29.28 &0.7974 &\textcolor{blue}{28.02} &0.7516 &27.89& 0.8336 &\textcolor{blue}{32.66} &0.9297\\
%%\rowcolor{blue!10}
MambaIRv2-L \cite{Guo2025MambaIRv2}& 4$\times$ & 34.2M & \textcolor{blue}{33.19} & \textcolor{blue}{0.9062} & \textcolor{blue}{29.29} & \textcolor{blue}{0.7982} & \textcolor{black}{28.01} & \textcolor{blue}{0.7521} & \textcolor{blue}{28.07} & \textcolor{blue}{0.8383} & \textcolor{blue}{32.66} & \textcolor{blue}{0.9304} \\
%%\rowcolor{red!10}
PS-Mamba-L (Ours) & 4$\times$ & 21.2M & \textcolor{red}{33.32} & \textcolor{red}{0.9076} & \textcolor{red}{29.41} & \textcolor{red}{0.8006} & \textcolor{red}{28.08} & \textcolor{red}{0.7537} & \textcolor{red}{28.27} & \textcolor{red}{0.8431} & \textcolor{red}{33.01} & \textcolor{red}{0.9324} \\
\bottomrule
\end{tabular}}
\label{tab:classic}
\end{table*}

% \vspace{-20}
\begin{table}[]
\centering
\caption{Quantitative comparison on different \textit{dimension sizes (D)} under the same training setting using Octant splitting.}
\setlength{\tabcolsep}{4pt}
\begin{tabular}{c|c|cc|cc|cc}
\hline
\multirow{2}{*}{D} & \multirow{2}{*}{\#Params} &
\multicolumn{2}{c|}{Set14} &
\multicolumn{2}{c|}{Urban100} &
\multicolumn{2}{c}{Manga109} \\ \cline{3-8}
 &  & PSNR & SSIM & PSNR & SSIM & PSNR & SSIM \\ \hline
24  & 277K & 34.08 & 0.9225 & 33.07 & 0.9375 & 39.44 & 0.9780 \\
48  & 970K & 34.38 & 0.9238 & 33.37 & 0.9394 & 39.74 & 0.9795 \\
64  & 1.666K & 34.45 & 0.9242 & 33.44 & 0.9399 & 39.81 & 0.9799 \\
%%\rowcolor{red!10}
96  & 3.597K & \textbf{34.51} & \textbf{0.9246} & \textbf{33.50} & \textbf{0.9404} & \textbf{39.87} & \textbf{0.9803} \\ \hline
\end{tabular}
\label{tab:dimension}
\end{table}

\vspace{-10pt}
\section{Experiments}
\label{sec:exper}
In this section, we evaluate the proposed method, Progressive Split-Mamba, across both \textit{lightweight} and \textit{classic} image super-resolution benchmarks, JPEG CAR (JPEG compression artifact reduction), denoising tasks (Gaussian color image denoising). All models are trained under similar settings for fair comparison, and results are reported on publicly adopted benchmark datasets: Set5, Set14, BSDS100, Urban100, and Manga109.
\subsection{Settings}
In order to maintain fair comparisons, we apply standard data augmentation techniques, including horizontal flipping and random rotations of 90$^\circ$, 180$^\circ$, and 270$^\circ$. During training, we extract 64~$\times$~64 patches for image SR and 128~$\times$~128 patches for image denoising. For the SR task,  the 2$\times$ model is first trained and then used to initialise the 3$\times$ and 4$\times$ models, while both the learning rate and total number of iterations are reduced by half to improve the final performance \cite{lim2017enhanced}. Also, for fair comparison, we set the batch size to 32 for SR and 8 for denoising and JPEG CAR. We use Adam~\cite{kingma2014adam} as the optimiser with $\beta_1=0.9$ and $\beta_2=0.999$.  We employ L1 loss for SR and Charbonnier loss for denoising and JPEG CAR. The initial learning rate is 2~$\times$~10$^{-4}$ and is reduced by half at predefined milestones. For SR, we provide two variants, lightweight and classic. Input images are padded to ensure the spatial dimensions are divisible by the required split factors used in the progressive partitioning strategy. This padding is removed after the reconstruction.
\subsection{Image Super-Resolution Comparisons}
\noindent\textbf{Lightweight Image Super-Resolution.}
Tables~\ref{tab:light} provides the quantitative comparison for $\times2$, $\times3$, and $\times4$ lightweight SR models. In line with prior studies~\cite{luo2020latticenet,srformer}, we additionally report the number of parameters (\#Params) and the multiply--accumulate operations (MACs), computed for upscaling a low-resolution image to a resolution of $1280 \times 720$, as measures of computational efficiency. PS-Mamba consistently outperforms existing lightweight methods such as CARN, LatticeNet, SwinIR-light, ELAN, and SRFormer-light, MambaIR, and MambaIRv2 by a clear margin. Across the various scales, PS-Mamba light establishes new state-of-the-art results in all datasets with the different varieties. For example, at $\times2$, it achieves 38.31~dB on Set5 and 33.37~dB on Urban100, improving upon SRFormer-light, MambaIR and MambaIRv2 by noticeable margins. Notably, the improvement is particularly significant on Set14, Urban100 and Manga109. This reflects that preserving the locality along with long-range dependencies is very crucial for better reconstruction. This also demonstrates the strength of the proposed progressive model of Mamba. Additionally, the comparison with the MambaIRv2 further highlights the competitiveness of our method. While MambaIRv2-light is a strong baseline, PS-Mamba-Light consistently achieves higher PSNR/SSIM across $\times2$, $\times3$, and $\times4$ under a comparable parameter budget. 

\begin{table}[t]
\centering
\caption{Ablation on the effectiveness of different split sizes.}
\setlength{\tabcolsep}{4pt}
\begin{tabular}{c|cc|cc|cc}
\hline
\multirow{2}{*}{Split Size} &
\multicolumn{2}{c|}{Set14} &
\multicolumn{2}{c|}{Urban100} &
\multicolumn{2}{c}{Manga109} \\ \cline{2-7}
 & PSNR & SSIM & PSNR & SSIM & PSNR & SSIM \\ \hline
Quarter     
& 34.28 & 0.9232 
& 33.27 & 0.9387 
& 39.64 & 0.9789\\
%%\rowcolor{red!10}
%
Octant      
& \textcolor{red}{34.38} & \textcolor{red}{0.9238} 
& \textcolor{red}{33.37} & \textcolor{red}{0.9394} 
& \textcolor{red}{39.74} & \textcolor{red}{ 0.9795} \\
%%\rowcolor{blue!10}
%
Sixteenth   
& \textcolor{blue}{34.28} & \textcolor{blue}{0.9232} 
& \textcolor{blue}{33.27} & \textcolor{blue}{0.9387} 
& \textcolor{blue}{39.64} & \textcolor{blue}{0.9789} \\ \hline
\end{tabular}
\label{tab:split_sizes}
\end{table}
\noindent\textbf{Classic Image Super-Resolution.}
In this part, we further evaluate PS-Mamba in the classic SR setting and compare with comparable Mamba-based methods, CNN-based approaches, attention-based methods including EDSR, RCAN, SAN, IPT, HAT, DAT, CAT-A, EDT, SwinIR, MaIR, and MambaIR. As shown in Tables~\ref{tab:classic}, our method in the three scales deliver consistent performance across all scales.
PS-Mamba achieves the best overall results; that is, outperforming prior Mamba- and transformer-based baselines. For instance, at $\times2$, it reaches 38.70~dB on Set5 and 40.90~dB on Manga109. This surpasses MambaIRV2 and all the previous methods with large margin. Our models achieve a better balance between restoration fidelity and model capacity. Notably, our model outperforms the largest model of MambaIRv2  with a clear margin. With 21.2M parameters, PS-Mamba-Large delivers consistent gains over MambaIRv2-L while using substantially fewer parameters (21.2M vs. 34.2M). Yet, with 13M of parameters less. For instance, it achieves a 0.20 dB PSNR improvement on the 4× Urban100 dataset. As illustrated in Fig. \ref{fig:four_images}, the visual results demonstrate that our method more effectively reconstructs sharp edges and realistic textures. Notably, PS-Mamba scales more efficiently, maintaining strong performance with fewer parameters, whereas MambaIRv2 requires substantially larger parameter growth. 
%

% \vspace{-12pt}
\subsection{Ablation Studies}
In this part, ablations with PS-Mamba light $\times 2$ SR model trained for 200,000 iterations on the three datasets, Set14, Urban100, and Manga109. This is an important study to show the influence of the new  components introduced in the method. We provided two ablation studies to analyse the performance of PS-Mamba under different channel dimensions and different split sizes.
\paragraph{Effect of Channel Dimension.}
Table~\ref{tab:dimension} shows the impact of the embedding dimension of the channels, while going down with the size and up on reconstruction quality. Increasing the channel dimension steadily improves performance. The best results obtained at dimension~96. This result aligns with prior findings that larger hidden states benefit state space models by enhancing reconstruction power. However, this setting is not preferable since the number of parameters increases rapidly, significantly reducing computational efficiency. Therefore, we adopt a fixed channel increment of 48, and all results are reported under this configuration.

%\vspace{-40pt}
\paragraph{Effect of Split Size.}
\begin{figure*}[!t]
\centering
\setlength{\tabcolsep}{0pt}     % remove horizontal padding
\renewcommand{\arraystretch}{1.5}  % vertical spacing between rows

\begin{tabular}{cc}
\includegraphics[width=0.5\textwidth, trim=0 4 0 4, clip]{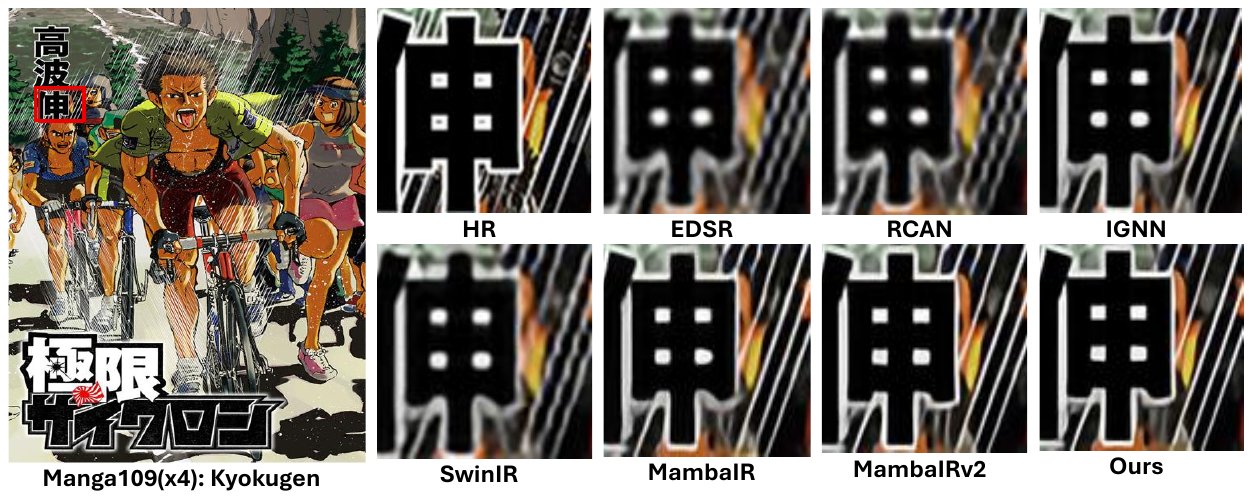}  &
\includegraphics[width=0.5\textwidth, trim=0 4 0 4, clip]{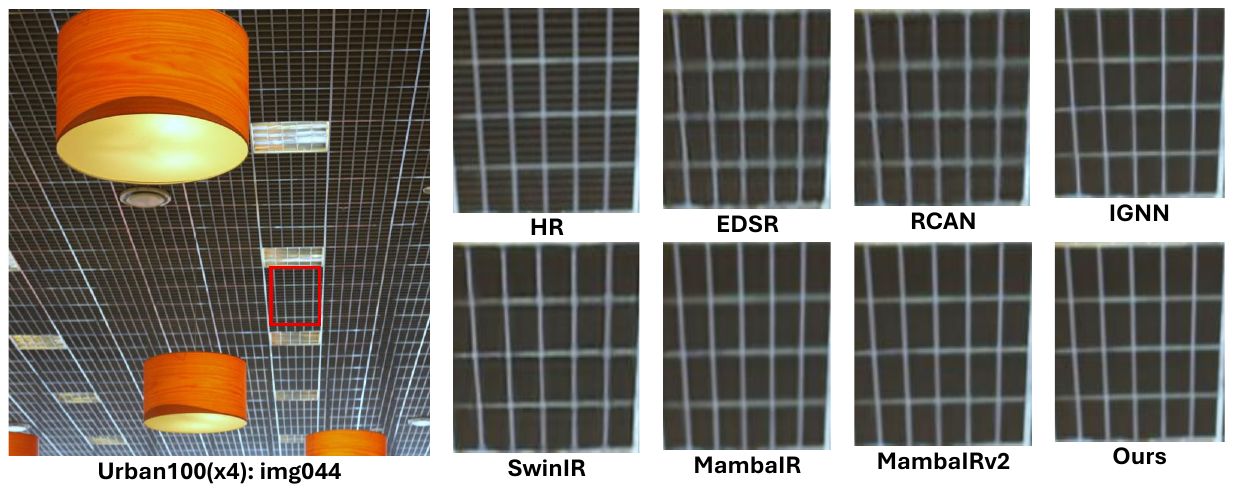}\vspace{-240pt} \\

\includegraphics[width=0.5\textwidth, trim=0 4 0 4, clip]{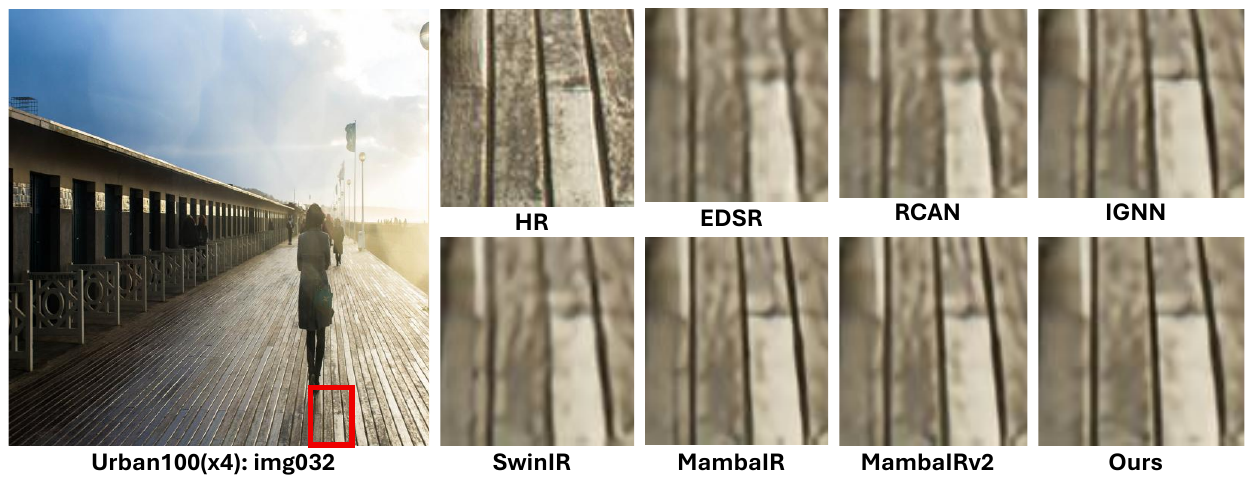} &
\includegraphics[width=0.5\textwidth, trim=0 4 0 4, clip]{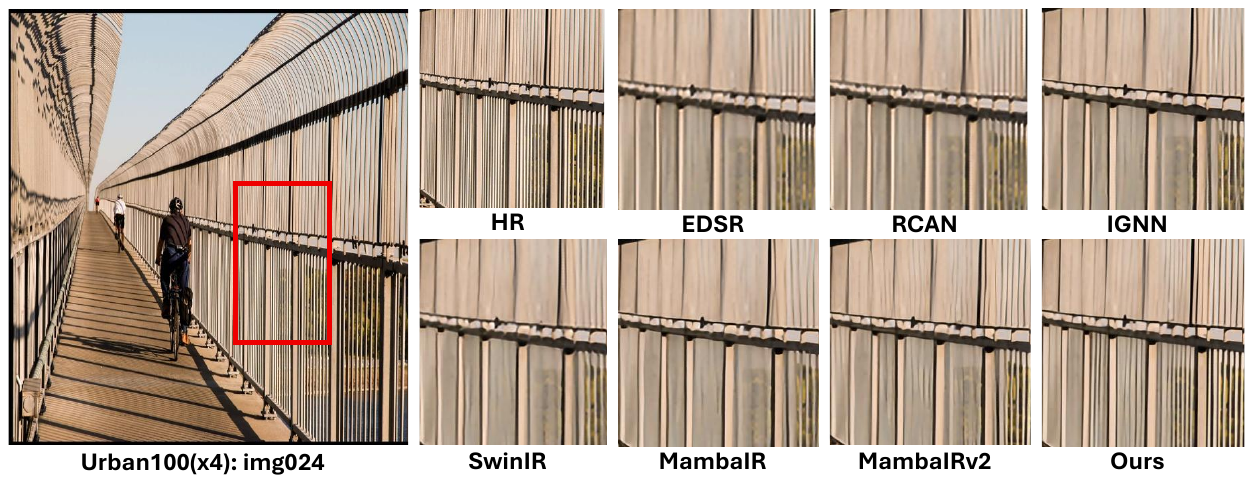} \\
\end{tabular}
\vspace{-210pt}
% \label{fig:visuals}
\caption{Visual comparison of HR , EDSR, RCAN, IGNN, SwinIR, MambaIR, MambaIRv2, and PS-Mamba on four picture from Manga109 and Urban100. PS-Mamba shows the closest to HR
}
\label{fig:four_images}
\end{figure*}
%\vspace{-21pt}
Table~\ref{tab:split_sizes} analyses the influence of different feature split sizes on performance. It is mainly about how many split are needed to converge the best performance. Using octant-level splitting yields the best results across all benchmarks. This confirms that a moderate number of partitions provides the optimal balance between local detail modelling and global dependency aggregation. Considering the size of the image, octants yield the best performance which balances between the minimum magnitude between the various tokens and the overall computation cost. However, splitting beyond octants does not yield further gains, likely because overly small regions weaken local continuity and introduce boundary effects as the results of Sixteenth's row reveal.
% \begin{table}[h]
% \centering
% \caption{Ablation on the effectiveness of different split sizes.}
% \setlength{\tabcolsep}{4pt}
% \begin{tabular}{c|cc|cc|cc}
% \hline
% \multirow{2}{*}{Split Size} &
% \multicolumn{2}{c|}{Set14} &
% \multicolumn{2}{c|}{Urban100} &
% \multicolumn{2}{c}{Manga109} \\ \cline{2-7}
%  & PSNR & SSIM & PSNR & SSIM & PSNR & SSIM \\ \hline
% %
% Quarter     
% & 37.80 & 0.9598 
% & 33.72 & 0.9189 
% & 32.10 & 0.8998 \\
% %%\rowcolor{red!10}
% %
% Octant      
% & \textcolor{red}{38.26} & \textcolor{red}{0.9614} 
% & \textcolor{red}{34.05} & \textcolor{red}{0.9214} 
% & \textcolor{red}{32.37} & \textcolor{red}{0.9021} \\
% %%\rowcolor{blue!10}
% %
% Sixteenth   
% & \textcolor{blue}{38.09} & \textcolor{blue}{0.9609} 
% & \textcolor{blue}{33.92} & \textcolor{blue}{0.9206} 
% & \textcolor{blue}{32.30} & \textcolor{blue}{0.9015} \\ \hline
% %
% \end{tabular}
% \label{tab:split_sizes}
% \end{table}
%

\subsection{Model Complexity Comparison.}
As shown in Table \ref{tab:complexity}, the proposed PS-Mamba-Large achieves strong performance while maintaining competitive computational complexity. With $21.2M$ parameters and $607.5G$ MACs, our model attains 34.71 dB / 0.9486 SSIM on Urban100 and 40.90 dB / 0.9814 SSIM on Manga109, outperforming several existing transformer-based approaches. Notably, PS-Mamba-Large achieves higher reconstruction accuracy than MambaIRv2-L, while using considerably fewer parameters (21.2M vs. 34.2M) and lower computational cost. These results demonstrate that PS-Mamba effectively balances restoration performance and efficiency, making it a powerful yet efficient backbone for image super-resolution.
\begin{table}[t]
\centering
\caption{Complexity comparison with state-of-the-art methods on 2$\times$ SR with output size $256\times256$.}
\setlength{\tabcolsep}{4pt}
\begin{tabular}{l|cc|cc|cc}
\hline
Models & \#Params & MACs & \multicolumn{2}{c|}{Urban100} & \multicolumn{2}{c}{Manga109} \\
 &  &  & PSNR & SSIM & PSNR & SSIM \\ \hline
CAT-A~\cite{cata2023} & 16.6M & 350.7G & 34.26 & 0.9440 & 40.10 & 0.9805 \\
DAT~\cite{dat2023} & 14.8M & 265.7G & 34.37 & 0.9458 & 40.33 & 0.9807 \\
HAT~\cite{hat2023} & 20.8M & 514.9G & 34.45 & 0.9466 & 40.26 & 0.9809 \\
% MambaIRv2-S~\cite{liu2024vmamba} & 9.6M & 192.9G & 34.24 & 0.9454 & 40.27 & 0.9808 \\
% MambaIRv2-B~\cite{liu2024vmamba} & 22.9M & 445.8G & 34.49 & 0.9468 & 40.42 & 0.9810 \\
MambaIRv2-L~\cite{liu2024vmamba} & 34.2M & 664.5G & 34.60 & 0.9475 & 40.55 & 0.9807 \\
PS-Mamba-Large (Ours) & \textbf{21.2M} & \textbf{607.5G} & 34.71 & 0.9486 & 40.90 & 0.9814 \\
\hline
\end{tabular}
\label{tab:complexity}
\end{table}
\vspace{-20pt}
\begin{table*}[!t]
\centering
\scriptsize
\caption{Quantitative comparison on \textit{JPEG compression artifact reduction} under different quality factors $q$.}
\resizebox{\textwidth}{!}{
\begin{tabular}{c|c|cc|cc|cc|cc|cc|cc}
\hline
\multirow{2}{*}{Dataset} & \multirow{2}{*}{$q$} 
& \multicolumn{2}{c|}{RNAN~\cite{zhang2019residual}} 
 
& \multicolumn{2}{c|}{DRUNet~\cite{zhang2021plug}} 
& \multicolumn{2}{c|}{SwinIR~\cite{liang2021swinir}} 
& \multicolumn{2}{c|}{MambaIR~\cite{Guo2024MambaIR}} 
& \multicolumn{2}{c|}{\cellcolor{blue!15}MambaIRv2 \cite{Guo2025MambaIRv2}} 
& \multicolumn{2}{c}{\cellcolor{red!15}\textbf{PS-Mamba (Ours)}} \\
& & PSNR & SSIM& PSNR & SSIM & PSNR & SSIM 
& PSNR & SSIM 
& \cellcolor{blue!10}PSNR & \cellcolor{blue!10}SSIM 
& \cellcolor{red!10}PSNR & \cellcolor{red!10}SSIM \\
\hline

% ------------ CLASSIC5 q10 ------------
\multirow{3}{*}{Classic5} 
& 10 
& 29.96 & 0.8178 

& 30.16 & 0.8234 
& 30.27 & 0.8249
& 30.27 & 0.8256
& \cellcolor{blue!10}\textcolor{blue}{30.37} & \cellcolor{blue!10}\textcolor{blue}{0.8269}
& \cellcolor{red!10}\textcolor{red}{\textbf{30.41}} & \cellcolor{red!10}\textcolor{red}{\textbf{0.8281}} \\

% ------------ CLASSIC5 q30 ------------
& 30 
& 33.38 & 0.8924 
 
& 33.59 & 0.8949 
& 33.73 & 0.8961 
& 33.74 & 0.8965
& \cellcolor{blue!10}\textcolor{blue}{33.81} & \cellcolor{blue!10}\textcolor{blue}{0.8970}
& \cellcolor{red!10}\textcolor{red}{\textbf{33.86}} & \cellcolor{red!10}\textcolor{red}{\textbf{0.8983}} \\

% ------------ CLASSIC5 q40 ------------
& 40 
& 34.27 & 0.9061 

& 34.41 & 0.9075 
& 34.52 & 0.9082 
& 34.53 & 0.9084
& \cellcolor{blue!10}\textcolor{blue}{34.64} & \cellcolor{blue!10}\textcolor{blue}{0.9093}
& \cellcolor{red!10}\textcolor{red}{\textbf{34.69}} & \cellcolor{red!10}\textcolor{red}{\textbf{0.9106}} \\
\hline

% ------------ LIVE1 q10 ------------
\multirow{3}{*}{LIVE1} 
& 10 
& 29.63 & 0.8239 
 
& 29.79 & 0.8278 
& 29.86 & 0.8287 
& 29.88 & 0.8301
& \cellcolor{blue!10}\textcolor{blue}{29.91} & \cellcolor{blue!10}\textcolor{blue}{0.8301}
& \cellcolor{red!10}\textcolor{red}{\textbf{29.95}} & \cellcolor{red!10}\textcolor{red}{\textbf{0.8313}} \\

% ------------ LIVE1 q30 ------------
& 30 
& 33.45 & 0.9149 

& 33.59 & 0.9166 
& 33.69 & 0.9174 
& 33.72 & 0.9179
& \cellcolor{blue!10}\textcolor{blue}{33.73} & \cellcolor{blue!10}\textcolor{blue}{0.9179}
& \cellcolor{red!10}\textcolor{red}{\textbf{33.78}} & \cellcolor{red!10}\textcolor{red}{\textbf{0.9192}} \\

% ------------ LIVE1 q40 ------------
& 40 
& 34.47 & 0.9299 
 
& 34.58 & 0.9312 
& 34.67 & 0.9317 
& 34.70 & 0.9320
& \cellcolor{blue!10}\textcolor{blue}{34.73} & \cellcolor{blue!10}\textcolor{blue}{0.9323}
& \cellcolor{red!10}\textcolor{red}{\textbf{34.78}} & \cellcolor{red!10}\textcolor{red}{\textbf{0.9336}} \\
\hline

\end{tabular}
}
\label{tab:jpeg_car}
\end{table*}
\subsection{Gaussian Colour Image Denoising}
One of the experiments that show the significance of PS-Mamba is evaluating on colour image denoising with noise level $\sigma = 15$. As shown in Table~\ref{tab:denoise}, the proposed method achieves the best performance on all datasets (CBSD68, Kodak24, McMaster, Urban100). This comparison to the previous state—of-the art including DRUNet, SwinIR, Restormer, Xformer, EAMambaIR MambaIR, and MambaIRv2, our model sets new benchmarks, reaching 34.55~dB on CBSD68 and 35.78~dB on McMaster. These results demonstrate that PS-Mamba acts as a strong and generalizable backbone for image restoration tasks, achieving superior denoising performance while maintaining a simple straight-through architecture, even when compared with U-shaped UNet-based models such as Restormer \cite{zamir2022restormer}, which are typically considered advantageous for denoising tasks.
\begin{table}[!tb]
\centering
\caption{Quantitative comparison of PSNR on \textit{gaussian color image denoising} $\sigma = 15$ with state-of-the-art methods.}
% \setlength{\tabcolsep}{4pt}
% \resizebox{\columnwidth}{!}{
\begin{tabular}{l|cccc}
\hline
\textbf{Method} & \textbf{CBSD68} & \textbf{Kodak24} & \textbf{McMaster} & \textbf{Urban100} \\ \hline
% IRCNN~\cite{zhang2017learning}    & 33.86 & 34.69 & 34.58 & 33.78 \\
% FFDNet~\cite{zhang2018ffdnet}  & 33.87 & 34.63 & 34.66 & 33.83 \\
DnCNN~\cite{zhang2017beyond}    & 33.90 & 34.60 & 33.45 & 32.98 \\
DRUNet~\cite{zhang2021plug}  & 34.30 & 35.31 & 35.40 & 34.81 \\
SwinIR~\cite{lim2017enhanced}  & 34.42 & 35.34 & 35.61 & 35.13 \\
Restormer~\cite{zamir2022restormer} & 34.40 & 35.35 & 35.61 & 35.13 \\
Xformer~\cite{zhang2023xformer} & 34.43 & 35.39 & 35.68 & 35.29 \\
MambaIR~\cite{Guo2024MambaIR} & {\color{blue}34.48} & 35.42 & 35.70 & 35.37 \\
EAMambaIR~\cite{lin2025eamamba} & 34.43 & 35.36 & 35.59 & -- \\
%%\rowcolor{blue!10}
MambaIRv2~\cite{Guo2025MambaIRv2} & {\color{blue}34.48} & {\color{blue}35.43} & {\color{blue}35.73} & {\color{blue}35.42} \\
%%\rowcolor{red!10}
\textbf{PS-Mamba (Ours)} & {\color{red}\textbf{34.55}} & {\color{red}\textbf{35.49}} & {\color{red}\textbf{35.78}} & {\color{red}\textbf{35.50}} \\ \hline
\end{tabular}
% }
\label{tab:denoise}
\end{table}

%\vspace{-10}
\subsection{Comparison on JPEG CAR.}
Table~\ref{tab:jpeg_car} illustrates the visual results on JPEG compression artifact reduction. 
Across all datasets and quality factors, our PS-Mamba  outperforms the compared methods including MambaIRv2. This confirms that the proposed method establishes in all datasets a new state-of the art. 
For instance, on the Classic5 dataset with $q=40$, PS-Mamba surpasses MambaIRv2 by 0.05\,dB in PSNR (34.69\,dB vs.\ 34.64\,dB). This reflects the strength ability to reconstruct structural details under compressed settings. 
Clearly, our PS-Mamba further boosts restoration reconstruction, as well as achieving the best results in every evaluation. These steady improvements across all quality factors illustrate the efficacy of PS-Mamba for JPEG artefact reduction.

%\vspace{-35pt}
\section{Conclusion}
\label{sec:conclusion}
We introduced \emph{Progressive Split-Mamba} (PS-Mamba), a state-space
restoration framework that resolves two core limitations of applying 1D
state-space models to 2D images: locality distortion and long-range decay. By
progressively splitting feature maps into topology-preserving patches, PS-Mamba
preserves spatial adjacent regions before state propagation and enables stable
modelling of fine textures. In parallel, the proposed symmetric cross-scale skip
propagation mitigates the exponential attenuation of long-range information,
which ensures global structural cues remained accessible throughout the network.

Combined with an adaptive Mamba–Conv block and dual-attention refinement,
PS-Mamba achieves a balanced framework that is capable of capturing both global
context and local details. Experiments verify consistent gains in structural
consistency, texture fidelity, and restoration accuracy. Overall, PS-Mamba
offers a simple yet effective methodology for stabilising Mamba-based 
models without token reordering or multi-directional scans, providing a
foundation for future extensions in image and video restoration. Despite its strong performance, PS-Mamba introduces additional split/merge operations that may increase implementation complexity. We will investigate more efficient implementations in future work.

\bibliographystyle{IEEEtran}
\bibliography{main}


\end{document}